\pdfoutput=1

\documentclass[11pt]{article}

\usepackage[review]{acl}

\usepackage{times}
\usepackage{latexsym}
\usepackage{graphicx}
\usepackage{lscape}
\usepackage{longtable}
\usepackage{makecell}
\usepackage{times}
\usepackage{booktabs}
\usepackage{array}
\usepackage{hyperref}

\usepackage[T1]{fontenc}

\usepackage[utf8]{inputenc}
\usepackage{microtype}

\title{The ADAIO System at the BEA-2023 Shared Task on Generating AI Teacher Responses in Educational Dialogues}

\author{Adaeze Adigwe \\
Center for Speech Technology Research\\ University of Edinburgh \\ United Kingdom \\
  \texttt{A.O.I.Adigwe@sms.ed.ac.uk} \\\And
  Zheng Yuan \\
  Istituto Italiano di Tecnologia,\\
  Università di Ferrara, \\ Italy \\
  \texttt{zheng.yuan@iit.it} \\}

\begin{document}
\maketitle
\begin{abstract}
This paper presents the ADAIO team's system entry in the Building Educational Applications (BEA) 2023 Shared Task on Generating AI Teacher Responses in Educational Dialogues. The task aims to assess the performance of state-of-the-art generative models as AI teachers in producing suitable responses within a student-teacher dialogue. Our system comprises evaluating various baseline models using OpenAI GPT-3 and designing diverse prompts to prompt the OpenAI models for teacher response generation. After the challenge, our system achieved second place by employing a few-shot prompt-based approach with the OpenAI \textit{text-davinci-003} model. The results highlight the few-shot learning capabilities of large-language models, particularly OpenAI's GPT-3, in the role of AI teachers.
\end{abstract}

\section{Introduction}
The current success of large language models (LLMs) in generating natural language responses that are almost indistinguishable from that of a human indicates that AI systems are steps closer to passing the Turing test. Apart from being used as conversational agents, LLMs can be employed in various educational settings as described in \citet{kasneci2023chatgpt} including as an AI teacher to help students practice and improve. \citet{tack_bea_2023} launches a shared task at the 18th Workshop on Innovative Use of NLP for Building Educational Applications (BEA 2023), called Generating AI Teacher Responses in Educational Dialogues. Inspired by \citet{tack2022ai}, this task requires teams to develop Intelligent Tutoring Systems (ITS) that generate teacher responses in real-world teacher-student interactions.  This task serves as a benchmark to gauge the capability of generative models in functioning as AI teachers.

Dialogue-based ITS face various requirements and challenges in meeting the needs of effective educational support. This entails generating factually accurate content and ensuring educational efficacy by speaking to students in a teacher-like manner, understanding their needs, and helping them improve their understanding \cite{tack2022ai}. However, several challenges must be addressed.

One significant challenge lies in acquiring appropriate data for training ITS, particularly real teacher-student interactions that cover various subjects. Another challenge involves developing models that can effectively capture the student's learning style and accommodate long-range dependencies within conversational sequences. Furthermore, evaluating the quality of teacher responses is essential. The responses should not only sound natural but also demonstrate an understanding of the student's queries and provide valuable guidance to help the student improve.



\section{Related Work}
Research on Intelligent Tutoring Systems has spanned many decades, with various proposed systems that include both text-based \cite{graesser2005autotutor}, spoken dialogue tutoring systems \cite{litman2004itspoke} and multi-modal systems that have been developed to improve student learning.  

Earlier dialogue-based ITS were designed using rule-based cognitive modelling methods \citep{aleven2010rule, vanlehn2002architecture} in generating teacher responses. In recent years natural language generation (NLG) tasks generally benefited from models using sequence-to-sequence architectures \cite{sutskever2014sequence}. Current state-of-the-art models such as OpenAI GPT-3 \cite{brown2020language} have shown tremendous results on a range of downstream NLG tasks such as response generation. One of the major underlying components of the language model is the transformer architecture \cite{vaswani2017attention} which increases its capacity for context awareness and long-range dependencies. Currently, the application of LLMs within the educational domain \cite{bibauw2022dialogue, hendrycks2021measuring} indicates they could improve student learning outcomes. However, their efficacy in conversational tutoring has not been fully evaluated \cite{tack2022ai}. 

On bench-marking the efficacy of LLMs in generating responses to accomplishing teaching goals, \citet{tack2022ai} investigate the suitability of these AI-teacher responses by comparing text generated by state-of-the-art models, Blender \cite{roller2020recipes} and GPT-3, on real-world tutoring dialogue data. The paper comparatively analyses the responses based on a stack of evaluation methods. Furthermore, the paper suggests the following  pedagogical dimensions to evaluate the AI-teacher generated responses, on its ability to \textit{speak like a teacher, understand a student and help a student}. These dimensions form the core of the AI-teacher challenge.

\section{Dataset}
\paragraph{Teacher-Student Chatroom Corpus (TSCC)} The dataset used in this task is derived from the Teacher-Student Chatroom Corpus (TSCC)\cite{caines2020teacher}. The TSCC consists of 102 chatrooms where English as a second language (ESL) teachers interact with students to work on language exercises and assess students' language proficiency. From each dialogue, shorter passages limited to 100 tokens were extracted, comprising sequential turns between the teacher and student. These passages serve as data samples and end with the teacher's utterance, which acts as the reference response. The dataset follows a JSON format, including fields such as id, utterances (dialogue context), and response (teacher's ending utterance).

The dataset includes a train set of 2,747 dialogues with an average of 3.9 turns per dialogue (±2.2, max=17). The dev set consists of 305 dialogues with an average of 4.0 turns (±2.2, max=16), while the test set comprises 273 dialogues with an average of 2.6 turns (±1.5, max=11). The response lengths in the train set range from 1 to 66 words, with an average of 9.1 words (±8.2) whereas the dev and test sets are without the response data.

\begin{figure*}[ht]
  \centering
  \includegraphics[scale=0.55]{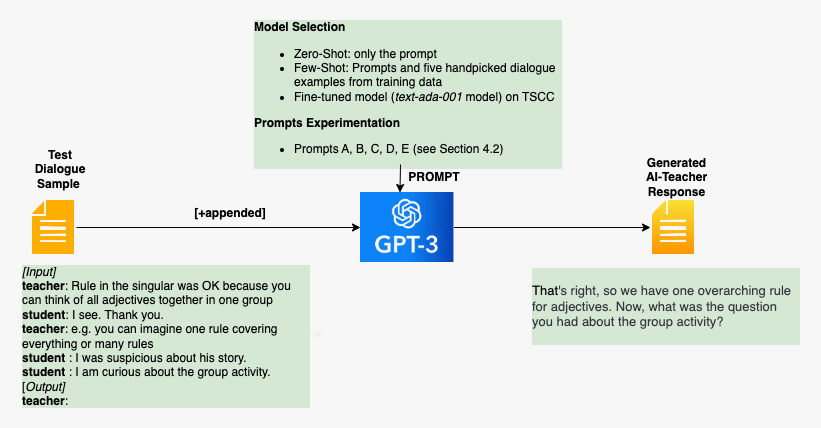} 
  \caption{The framework for the proposed GPT-3 based Intelligent Turing System. Depending on the experimental setup, the specified prompt followed by a few handpicked dialogue examples (if applicable) is sent to the LLM (GPT-3) to generate an AI-teacher response.}
  \label{fig:pipeline}
\end{figure*} 

\section{System Architecture}
\label{sub:sys}
\subsection{Model} 
\label{subsec:model}
We conducted our experiments using OpenAI GPT-3 \cite{brown2020language} pre-trained LLMs. Initial trials revealed that the \textit{text-davinci-003} model produced responses that closely resembled human-like and contextually relevant interactions, surpassing the performance of \textit{ada}, \textit{curie}, and \textit{babbage}. Consequently, we predominantly employed this model for our experiments. However, considering the cost associated with utilizing the models, we opted for the \textit{text-ada-001} model for the fine-tuning setting described below. A schematic overview of our experimental process is depicted in Figure~\ref{fig:pipeline}.

\subsection{Training Methods}
\label{subsec:train}
Earlier deep-learning models would employ fine-tuning techniques to update the parameters of a pre-trained model by retraining it on new data samples from the target domain. Pre-trained LLMs such as GPT-3 and others have demonstrated the ability to utilize natural language prompts either with or without accompanying examples in performing downstream NLP tasks such as classification, summarization or generation \cite{brown2020language, liu2023pre}. Within dialogue generation, fine-tuning with example data can lead to responses generated with desirable attributes or tones such as empathy, persuasion, encouragement, etc. In tutoring situations, there are attributes that make a good teacher, and we wanted to examine the ability of dialogue-based ITS to embody such characteristics.
The training methods we explored include zero-shot, few-shot, and fine-tuning settings.

\begin{enumerate}

    \item \textbf{Zero-Shot}: In this approach, we simply provided the GPT-3 Model with a modified version of \textbf{Prompt A} \textit{(see Section \ref{subsec:prompt})}, without any example dialogues.  
    
    \item \textbf{Few-shot}: This approach features adding to the prompts five handpicked sample dialogues (see \textit{Table} \ref{tab:dial_ex}) from the training set. These dialogues included the \textit{speaker-role} for each turn, i.e. \textit{student}, and \textit{teacher} just like in the training data. Our criteria were to choose dialogue examples with a teaching focus as defined in \citet{caines2020teacher}. As per the teaching focus, we selected example dialogues that consisted of conversational sequences that sought to provide grammatical and lexical resources to the student while also showing aspects of discourse management and interactive communication. We replicated this approach using two language models, namely \textit{text-ada-001} and \textit{text-davinci-003}.

    \item \textbf{Fine-tuning on the TSCC corpus}: We fine-tuned the \textit{text-ada-001} model on the training data following OpenAI's API documentation \href{https://platform.openai.com/docs/guides/fine-tuning}{(https://platform.openai.com/docs/guides/fine-tuning)}. Our fine-tuned data consisted of approximately 95$\%$ of the training data, excluding the test data that we set aside for our internal evaluation. Afterwards, we used the fine-turned data to prompt the model, exactly like the few-shot approach to generate the teacher responses for the test sample. 
\end{enumerate}

\subsection{Prompts Engineering}
\label{subsec:prompt}
In this section, we delve into the adaptability of the dialogue-based Intelligent Tutoring System (ITS) by employing prompts that experiment with various aspects, including the roles of the participants, the teaching approach adopted by the tutor, and the specific teaching goals. To achieve this, we utilized the few-shot approach, providing explicit instructions to the model regarding dialogue response generation. The prompts used, along with corresponding dialogue examples, are presented below and in Table \ref{tab:dial_ex}.

 \begin{enumerate}
     \item \textbf{Prompt A} You will be given a dialogue chat between \textit{a teacher and a student}, and your task is to generate a teacher response that is appropriate to the context, in which the teacher is \textit{polite, helpful, professional, on topic, and factually correct}. The following are example dialogues with a teacher and a student.
     \item \textbf{Prompt B} You will be given a dialogue chat between \textit{a teacher and a student}, and your task is to generate a teacher response and \textit{probe the student's understanding} in a \textit{strict} manner. The following are example dialogues with a teacher and a student. 
     \item \textbf{Prompt C} You will be given a dialogue chat and your task is to  generate a teacher response. The following are example dialogues with a teacher and a student. 
   \item \textbf{Prompt D} You will be given a dialogue chat between an \textit{English language learner and a teacher}. Your task is to generate the teacher's response to \textit{encourage conversational skills}. The following are example dialogues with a teacher and a student. 
   \item \textbf{Prompt E} You will be given a dialogue chat between \textit{two conversational partners}. Generate the utterance that is appropriate within the dialogue context. The following are \textit{example dialogues}. 
 \end{enumerate}

Prompts A and B are designed to incorporate aspects of the tutor's teaching approach, with prompt A, exhibiting more desirable attributes (adopted from \citet{tack2022ai}). In contrast, prompt B adopts a slightly different approach to probe the learner’s understanding. Prompt C takes a neutral stance without any characteristics, putting more focus on the student-teacher roles of the dialogue participants. Prompt D attempts to generate responses that ought to focus on the learning goal - second language acquisition skills as specified in the TSCC corpus. Lastly, Prompt E removes the teacher-student roles and shifts towards dialogue participants with unspecified roles. The role tags in the few-shot examples are changed to Speaker A and Speaker B in this prompt. 

\subsection{Implementation Details}
We used the OpenAI Python library to call the GPT-3 engine to make the inferences on the test dialogues. Among the available models, we employed the top-performing \textit{text-davinci-003} in the zero-shot and few-shot scenarios, and \textit{text-ada-001} in the fine-tuned approach. Additionally, we compared the performance of \textit{davinci} and \textit{ada} in the few-shot experiments. We used the following parameters for all our experiments: \textit{temperature=0.7}, \textit{max tokens=100}, \textit{top p=0.8}, \textit{frequency penalty=0} and \textit{presence penalty=0}. We experimented with a range of values for \textit{max tokens} including \textit{20, 30, 70, 100, 256}. After some initial trials, we decided to go with \textit{max-tokens=100} as it generated both a concise and relevant response most of the time. Across all our trials we kept the parameters and settings the same. 

In our few-shot experimental settings, we intentionally disregarded examples samples that lacked teaching material in the reference teaching response such as turns that expressed acknowledgement, greetings or parenthetical statements, for example, conversational turns like \textit{sure, okay, hi, etc.}
The few-shot prompts dialogue examples were kept the same across the experiments. 

\section{Results}
\subsection{Model Selection}
We randomly selected fifty samples from the training data to constitute our \textit{internal test set} for model selection. These samples did not overlap with the few-shot dialogue examples and thus allowed us to compare the training methods listed in Section \ref{subsec:train}. We utilized the machine-based evaluation metric BertScore \cite{zhang2019bertscore} and reported the recall, precision, and F1 scores in Table ~\ref{tab:automaticevaluation_results} (all models were fed with Prompt A). The BERTScores show little variability across the models despite the apparent differences we noticed when inspecting the generated responses. Eventually, for our final system entry, we chose the Few-Shot \textit{davinci-003} model based on Prompt A as we believe this system generated the most meaningful responses required by the shared task. From our observation, both the few-shot and fine-tuned \textit{ada-001 models} generated out-of-context and incoherent responses most of the time. We abstain from reporting the BertScores of models fed by Prompt B to E for the performances were consistent as shown in Table \ref{tab:automaticevaluation_results} and that we didn't have the resources to engage human evaluation on the quality of the generated response. Nevertheless, the generated responses piqued our interest, leading us to incorporate a few in the Appendix.




\begin{table}[t]
  \centering
  \resizebox{\columnwidth}{!}{%
  \begin{tabular}{|l|l|l|l|}
    \hline
    \textbf{Models} & \textbf{Prec.} & \textbf{Rec.} & \textbf{F1} \\
    \hline
    Zero-Shot \textit{(davinci-003)} & 0.83 & \textbf{0.847} & 0.842 \\
    \hline
    Few-Shot \textit{(ada-001)} & \textbf{0.848} & 0.839 & \textbf{0.844} \\
    \hline
    Few-Shot \textit{(davinci-003)} & 0.840 & 0.844 & 0.842 \\
    \hline
    Finetuned \textit{(ada-001)} & 0.811 & 0.836 & 0.824 \\
    \hline
  \end{tabular}%
  }
  \caption{BertScore evaluation of models on the internal test set}
  \label{tab:automaticevaluation_results}
\end{table}

\begin{table*}[ht]
  \centering
  \resizebox{450px}{!}{%
  \begin{tabular}{|l|l|l|l|l|l|l|l|l|}
    \hline
    \multicolumn{1}{|c|}{} &
    \multicolumn{3}{|c|}{\textbf{BERTScore}} & \multicolumn{5}{|c|}{\textbf{DialogRPT}} \\
    \hline
    \textbf{ Phase} & \textbf{Prec.} & \textbf{Rec.} & \textbf{F1} 
    & \textbf{Updown} & \textbf{Human vs. Rand} 
    & \textbf{Human vs. Machine} & \textbf{Final (avg)} & \textbf{Final (best)}
    \\
    \hline
    \textit DEV Phase & 0.67(5) & 0.71(1) & \textbf{0.69(1)} 
    & 0.37(5) & 0.98(1) & 0.99(4) & \textbf{0.35(2)} & 0.71(6)
    \\
    \hline
    \textit EVAL Phase & 0.72(4) & 0.69(3) & \textbf{0.71(3)} 
    & 0.40(5) & 0.97(2) & 0.98(5) & \textbf{0.37(3)} & 0.65(7)
    \\
    \hline
  \end{tabular}%
  }
  \caption{BEA Shared Task official results of the \textit{adaio} system}
  \label{tab:BAE_res}
\end{table*}

\subsection{Shared Task Results}
\label{subsec:task_res}
Table \ref{tab:BAE_res} presents the results of our ADAIO System (Few-Shot \textit{davinci-003} model based on Prompt A) during the development and evaluation phases of the shared task. The numbers in parentheses represent the system's rank among the top 10 entries. The BertScore deviation observed as compared to the model selection results may be attributed to the variation in data between the reference responses in the \textit{real test set} and the training set. Apart from BertScore, the shared task incorporates another automated dialogue evaluation metric known as DialogRPT \cite{gao2020dialogrpt}. This metric assesses the generated response's performance in relation to the preceding dialogue context, considering indicators such as \textit{updown} (the average likelihood that the response receives the most upvotes), \textit{human vs rand} (the average likelihood that the response is contextually relevant), \textit{human vs machine} (the average likelihood that the response is human-written rather than machine-generated), and \textit{final} (the average/maximum) weighted ensemble score derived from all DialogRPT metrics. Our ADAIO System ranked second place after the two phases.

\section{Discussion}
The evaluation results from both machine and human assessments of the generated responses on the test set provide evidence of the effectiveness of LLMs, particularly GPT-3, in tutoring dialogue applications. While the dialogues in the TSCC corpus primarily concentrate on everyday speech and language usage, which proves advantageous for short conversational exchanges such as corrections, explanations, or clarifications, it is crucial to examine the GPT-3 model's reliability in tutoring scenarios that involve longer sequences within a wider discourse context \cite{graesser1995collaborative}. Furthermore, we perceived a limitation in relying solely on automatic evaluation metrics (as detailed in Section 5.1)

\textbf{Prompt engineering to adapt language and tone in tutoring systems} Our experiments reveal an intriguing finding where manipulating the prompt influences the tone and language of the generated response, presenting an opportunity for tutoring systems to potentially adapt to the students' learning styles and/or teaching goals. Further research should delve into teaching instruction methods, potentially exploring the pedagogy of constructivist learning \cite{graesser2005autotutor} or engaging students in ill-structured exercises for productive failure \cite{kapur2008productive} using LLMs of this nature. 

\textbf{GPT-3's robust handling of errors and non-canonical form of language} During the data preparation phase, a manual inspection of the data revealed the presence of grammatical and spelling errors in some utterances. Additionally, since the dataset originated from chatroom text-based conversations, there were instances where mathematical symbols were used instead of natural language, such as this example utterance \textit{Output teacher: But e.g. pleased with their visit = good idea}. It is worth noting that we did not employ any NLP processing toolkit to correct these errors or non-canonical forms in the dialogue utterances. However, despite this, the GPT-3 model could still generate appropriate responses effectively.

\textbf{LLMs' potential in multilingual settings} In the context of L2 acquisition, the dialogue nature in \citet{caines2020teacher} provides valuable opportunities for tutors to adapt to students' native languages. Code-switching strategies as such have been found to enhance teaching, including the explanation of concepts \cite{koppe199513}, and leveraging AI tutoring systems can facilitate this process. LLMs possess multilingual capabilities that enable them to address language barriers, accommodate low-resource languages, and exhibit promising performance even on unseen languages \cite{yong2022bloom+}. To enhance accessibility, the development and adoption of open-source multilingual models, such as BLOOM \cite{scao2022bloom}, should be encouraged, thereby facilitating the utilization of LLMs in educational applications across diverse linguistic contexts.

\section{Conclusion}
In this paper, we have presented our system entry to the BEA 2023 Shared Tasks on AI-teacher response generation. Our approach investigates the capability of the state-of-the-art language generative model, OpenAI GPT-3, in addressing the requirements of the AI teacher challenge outlined by \citealt{tack2022ai}. Through extensive experimentation utilizing zero-shot, few-shot, and fine-tuning techniques, we investigated the adaptability of the system's responses by leveraging meticulously designed prompts and carefully selected dialogue examples that emphasize desirable teacher qualities. Our submitted system, featuring a few-shot prompt-based method, achieved 2nd place in the BEA Shared Task 2023 challenge.

\section{Acknowledgements}
 The first and second authors have received funding
from the European Union’s Horizon 2020 research and
innovation program under the Marie Skłodowska Curie
grant agreement No 859588. The authors are also indebted to Carol Figueroa for the helpful comments and feedback on the paper revision. 

\bibliography{anthology,custom}
\bibliographystyle{acl_natbib}

\clearpage
\onecolumn
\appendix

\section{Appendix} \label{sec:appendix}
\begin{longtable}{|p{2cm}|p{12cm}|}
\caption{Sample dialogues and responses in the few-shot experiment} \label{tab:dial_ex} \\
\hline
\textbf{Field} & \textbf{Content} \\
\hline
\endfirsthead
\multicolumn{2}{c}{\textbf{Table continued from previous page}} \\
\hline
\textbf{Field} & \textbf{Content} \\
\hline
\endhead
\hline
\multicolumn{2}{r}{\textbf{Continued on next page}} \\
\endfoot
\hline
\endlastfoot

\hline
\hline
Dialogue ID & 0701 \\
\hline

\multicolumn{1}{|l|}{History} &
\multicolumn{1}{p{12cm}|}
{\begin{tabular}[c]{@{}p{12cm}@{}}
student: ok,\\ student: people who visited the museum after its regeneration were more pleased than the visitors in the years before\\ student: I'm not sure if I can use regeneration like a synonym \\ teacher: Um...good idea but not quite right here I'm afraid....maybe for e.g an atea of a city or a whe industry...\\ teacher: Sorry...area of a city... or whole industry\\ student: yes I see \\ student: because it's the same thing in Italian
\end{tabular}} \\
\hline

\multicolumn{1}{|l|}{Ref. Responses} &
\multicolumn{1}{p{12cm}|}{
teacher: But e.g. pleased with their visit = good idea "} \\
\hline
\multicolumn{1}{|l|}{Prompt A} &
\multicolumn{1}{p{12cm}|}{teacher:' Ah yes, "regeneration" isn't quite the right word here. We want to focus on the museum specifically, so it's better to use "renovation" instead.} \\
\hline
\multicolumn{1}{|l|}{Prompt B} &
\multicolumn{1}{p{12cm}|}{teacher: Yes, that's correct, but in this context you need to use the word 'renovation' instead of 'regeneration'. Can you explain why this is the case?} \\
\hline
\multicolumn{1}{|l|}{Prompt C} &
\multicolumn{1}{p{12cm}|}{teacher: Yes, you can use the term "regeneration" to refer to a broader concept, such as the transformation of a whole area or industry. But in this context, it's referring specifically to the renovation of the museum.} \\
\hline
\multicolumn{1}{|l|}{Prompt D} &
\multicolumn{1}{p{12cm}|}{teacher: Yes, that's right. You can use 'regeneration' as a synonym for 'renovation' in this context, but not for larger-scale changes like an area of a city or a whole industry.} \\
\hline
Prompt E & SpeakerB: So, what would be a better word to use here? \\
\hline

\multicolumn{2}{c}{} \\
\hline
\hline
Dialogue ID & 1062 \\
\hline
\multicolumn{1}{|l|}{History} &
\multicolumn{1}{p{12cm}|}
{\begin{tabular}[c]{@{}p{12cm}@{}}
student: Natural gas account for 20\% of the electricity in Australia in 1980. \\
teacher: yes that's great! What verb tense do you need?\\
student: Natural gas made up 20\% of electricity in Australia in 1980.\\
teacher: Yes perfect! Good past simple too\\
teacher: Account for in the past = accounted for\\
student: so...Natural gas accounted for 20 \% of electricity in Australia in 1980.\\
student: ok..thanks\\
student: I got it\\
\end{tabular}} \\
\hline

\multicolumn{1}{|l|}{Ref. Responses} &
\multicolumn{1}{p{12cm}|}{
teacher: Yes that's great - that would look good in an answer} \\
\hline

\multicolumn{1}{|l|}{Prompt A} &
\multicolumn{1}{p{12cm}|}{
teacher:' Great, that\'s exactly right! Remember, you can also use other phrases like "made up" or "constituted" to mean the same thing.
} \\
\hline

\multicolumn{1}{|l|}{Prompt B} &
\multicolumn{1}{p{12cm}|}{
teacher: Great, can you explain what this sentence means in your own words?
} \\
\hline

\multicolumn{1}{|l|}{Prompt C} &
\multicolumn{1}{p{12cm}|}{
teacher: Perfect! That's it.} \\
\hline

\multicolumn{1}{|l|}{
Prompt D} &
\multicolumn{1}{p{12cm}|}{teacher: Yes, that's correct! Great job!} \\
\hline

Prompt E & SpeakerB: Great, now you try it with the other numbers in the chart.  \\
\hline

\pagebreak
\hline
\hline
Dialogue ID & 0298 \\
\hline

\multicolumn{1}{|l|}{History} &
\multicolumn{1}{p{12cm}|}{\begin{tabular}[c]{@{}p{12cm}@{}}
teacher: It's a bit random to be honest...if we add certain words then sometimes the singular becomes more likely e.g. the fire service is...\\
teacher: Yes we often say the government are... \\
teacher: a bit less formal but common yes...like we're thinking of them as a bunch of people \\
student: ohh I didn't know that... \\
student: But I kind of get it haha \\
teacher: No I can see it might seem a bit strange \\
teacher: By the way I think the plural form is less common the USA \\
\end{tabular}} \\
\hline

\multicolumn{1}{|l|}{Ref. Responses} &
\multicolumn{1}{p{12cm}|}{
teacher: in the US
} \\
\hline

\multicolumn{1}{|l|}{Prompt A} &
\multicolumn{1}{p{12cm}|}{
teacher: That's great! Just remember that when talking about the government in the US, it's more common to use the singular form ("the government is") than the plural form ("the government are").
} \\
\hline

\multicolumn{1}{|l|}{Prompt B} &
\multicolumn{1}{p{12cm}|}{
teacher: That's great to know! Can you explain to me why the plural form is less common in the US than in Britain?
} \\
\hline

\multicolumn{1}{|l|}{Prompt C} &
\multicolumn{1}{p{12cm}|}{
teacher: Yes that's a good idea. It's important to be aware of the differences between different varieties of English.
} \\
\hline

\multicolumn{1}{|l|}{Prompt D} &
\multicolumn{1}{p{12cm}|}{
teacher: Yes, it can be tricky! I think it's important to note that the plural form is more commonly used in Britain than in the US. But don't worry, with practice you'll get the hang of it.
} \\
\hline

Prompt E & SpeakerB:  Yeah, definitely. Just remember to use "are" when talking about a group of people in the US, and "is" when talking about a single entity.\\
\hline

\multicolumn{2}{c}{} \\
\hline
\hline
Dialogue ID & 1674 \\
\hline
\multicolumn{1}{|l|}{History} &
\multicolumn{1}{p{12cm}|}{\begin{tabular}[c]{@{}p{12cm}@{}}
student: plug it out \\
teacher: nope! unplug it! \\
teacher: A plug is also the bit at the end of a cable that you put into an electric socket \\
teacher: Plugs can look different in different countries. For example, in Ukraine, a plug has two pins \\
teacher: Do you know what I'm talking about? \\
student: yes \\
\end{tabular}} \\
\hline

\multicolumn{1}{|l|}{Ref. Responses} &
\multicolumn{1}{p{12cm}|}{
teacher: Ok great! So you know that if you want to turn something off you unplug it, right? 
} \\
\hline

\multicolumn{1}{|l|}{Prompt A} &
\multicolumn{1}{p{12cm}|}{
teacher: Great! So, what do you need to do with the plug?'
} \\
\hline

\multicolumn{1}{|l|}{Prompt B} &
\multicolumn{1}{p{12cm}|}{
teacher: Great! Can you explain to me what a plug is and how it works?
} \\
\hline

\multicolumn{1}{|l|}{Prompt C} &
\multicolumn{1}{p{12cm}|}{
teacher: Great! So when you need to unplug something, make sure you do it carefully and don't pull on the cable itself
} \\
\hline

\multicolumn{1}{|l|}{Prompt D} &
\multicolumn{1}{p{12cm}|}{
teacher: Great, can you explain what kind of plugs are used in other countries?
} \\
\hline

Prompt E & SpeakerB: So what do you need to do with it? Unplug it or plug it in?\\
\hline

\multicolumn{2}{c}{} \\
\hline
\hline
Dialogue ID & 1650 \\
\hline

\multicolumn{1}{|l|}{History} &
\multicolumn{1}{p{12cm}|}{\begin{tabular}[c]{@{}p{12cm}@{}}
student: informally you hear people say e.g. So I go to the yoga class and it's so hard etc Do that people mean that they go everyday or everyweek and suffer? \\
teacher: By the way, the structure:  'I changed my idea' is a bit weird...normal - e.g. it (the experience) changed my idea \\
student: I changed my mind would sound better??
\end{tabular}} \\
\hline

\multicolumn{1}{|l|}{Ref. Responses} &
\multicolumn{1}{p{12cm}|}{
teacher: Ah well that would be normal in that meaning but no - i mean to describe a specific past situation
} \\
\hline

\multicolumn{1}{|l|}{Prompt A} &
\multicolumn{1}{p{12cm}|}{
teacher: Yes, "I changed my mind" would be a better way of expressing it. As for the yoga class, it could mean that they attend the class on a regular basis and find it challenging.
} \\
\hline

\multicolumn{1}{|l|}{Prompt B} &
\multicolumn{1}{p{12cm}|}{
teacher: Yes, "I changed my mind" would sound better. To answer your question, it depends on the context. It could mean that they go to the yoga class regularly and find it challenging, or it could mean that they went once and found it difficult.
} \\
\hline

\multicolumn{1}{|l|}{Prompt C} &
\multicolumn{1}{p{12cm}|}{
teacher:  Yes, "I changed my mind" would sound better. As for the yoga class, it depends on what the person is saying - it could mean they go every day, or it could mean they go occasionally and find it difficult.'
} \\
\hline

\multicolumn{1}{|l|}{Prompt D} &
\multicolumn{1}{p{12cm}|}{
Yes, "I changed my mind" would sound better. It\'s important to use the right grammar when speaking English. But, to answer your question, it depends on the context of the conversation. People might mean that they go to the yoga class every day or every week and find it difficult, or they might just mean that it was difficult that particular time.
} \\
\hline

Prompt E & SpeakerB:  Yes, 'I changed my mind' would sound better :)\\
\hline

\end{longtable}

\end{document}